\documentclass[a4]{article}

\usepackage{authblk}
\usepackage[cmex10]{amsmath}
\usepackage{amsfonts}
\usepackage{url}
\usepackage{hyperref}

\newcommand{\argmin}{\operatornamewithlimits{argmin}}

\begin{document}

\title{MEG Decoding Across Subjects}

\author{Emanuele Olivetti, Seyed Mostafa Kia and Paolo Avesani}

\affil{NeuroInformatics Laboratory (NILab),\\
  Bruno Kessler Foundation, Trento, Italy\\
  Centro Interdipartimentale Mente e Cervello (CIMeC),\\
  University of Trento, Italy
}

\maketitle

\begin{abstract}
  Brain decoding is a data analysis paradigm for neuroimaging
  experiments that is based on predicting the stimulus presented to
  the subject from the concurrent brain activity. In order to make
  inference at the group level, a straightforward but sometimes
  unsuccessful approach is to train a classifier on the trials of a
  group of subjects and then to test it on unseen trials from new
  subjects. The extreme difficulty is related to the structural and
  functional variability across the subjects. We call this approach
  \emph{decoding across subjects}. In this work, we address the
  problem of decoding across subjects for magnetoencephalographic
  (MEG) experiments and we provide the following contributions: first,
  we formally describe the problem and show that it belongs to a
  machine learning sub-field called transductive transfer learning
  (TTL). Second, we propose to use a simple TTL technique that
  accounts for the differences between train data and test
  data. Third, we propose the use of ensemble learning, and
  specifically of stacked generalization, to address the variability
  across subjects within train data, with the aim of producing more
  stable classifiers. On a face vs. scramble task MEG dataset of 16
  subjects, we compare the standard approach of not modelling the
  differences across subjects, to the proposed one of combining TTL
  and ensemble learning. We show that the proposed approach is
  consistently more accurate than the standard one.
\end{abstract}


\section{Introduction}
\label{sec:introduction}

Predicting the mental state of a subject from concurrent neuroimaging
data is a data analysis approach usually called \emph{brain
  decoding}. The subject is presented a stimulus, e.g. a visual cue,
and the related brain activity is recorded from multiple sensors. The
recorded data, together with the category of the stimulus, are denoted
as \emph{trial}. During an experiment multiple trials are collected
and in the brain decoding analysis phase, part of the data, i.e. the
\emph{train set}, is used to build a classifier, i.e. a function that
predicts the stimulus from the brain activity. The remaining part of
the data are used to estimate how accurate the classifier is, i.e. to
test its generalisation ability \cite{lemm2011introduction}. Accurate
classification is considered evidence that stimulus-related
information is present in the data, a conclusion that shed light on
the relation between the mental process of interest and its neural
correlates. In the following, we refer to the case
magnetoencephalographic (MEG) data and defer the general case of
functional neuroimaging data to future work.

Brain decoding is usually performed at the level of each single
subject and inference at the group-level is frequently conceived as
the analysis of the set of the single subject results, e.g. a
one-sample $t$-test over the set of single-subject accuracies
(See~\cite{olivetti2011bayesian} for a critical discussion). Ideally,
training a classifier on the trials from a set of subjects and testing
the classifier on trials from other unseen subjects is a meaningful
process to make inference at the group level. Unfortunately this
approach is technically difficult because of the structural and
functional differences of the brain across subjects together with the
inherent variability of the MEG measurements due, for example, to
changes in environmental variables. The effect of this variability is
that the underlying probability distribution of the trials changes
from subject to subject violating the assumption of a single
underlying generative process~\footnote{The specific assumption is
  that, for each category of stimulus, the concurrent neuroimaging
  data are \emph{independently and identically distributed} (i.i.d.).}
between the train set and the test set, which is the cornerstone of
statistical learning and, in general, of inference. In practical
terms, it is common experience to observe that the average accuracy of
the classifiers trained and tested on each individual subject is much
greater than that of a classifier trained on pooling trials of a set
of subjects and tested on trials from other subjects. We provide
empirical evidence of this fact in Section~\ref{sec:experiments}.

To the best of our knowledge, the literature proposing solutions to
the problem of MEG decoding across subjects for inferential purpose is
at very early stage. Most of the effort is in creating a
representation of the MEG data, i.e. a new feature space, that is as
homogeneous as possible across subjects. A recent example is
in~\cite{kauppi2013decoding}.

Other fields of research are related, to some extent, to the problem
of MEG decoding across subjects. They can be divided into three main
different areas:
\begin{itemize}
\item Machine Learning literature dealing with training and testing on
  different feature spaces and/or different distributions - a general
  paradigm known as \emph{transfer learning}. See~\cite{pan2010survey}
  for a recent review. To the best of out knowledge, in this community
  no applications have been presented in the context of MEG data
  analysis.
\item Brain-computer interface (BCI) literature on between-subject
  learning from EEG data (see for
  example~\cite{devlaminck2011multisubject}). In this setting, data
  from other subjects is used to improve the calibration of BCI
  devices when used on a new subject. The stimulus information is
  required for at least some trials of the new subject. This
  requirement is different from our setting where no stimulus
  information is available on test subjects. Notice that this
  technical difference leads to a completely different approach during
  data analysis.
\item Literature about decoding across subjects from functional
  magnetic resonance imaging (fMRI) data. To the best of our
  knowledge, the two leading directions are \emph{hyperalignement}
  (see for example~\cite{haxby2011common}) and multi-task learning
  (see for example~\cite{marquand2014bayesian}). Both differ from our
  setting: hyperalignement requires an additional recording from all
  subjects with identical rich stimulation. In multi-task learning it
  is assumed that some stimulus information is available on test
  data~\footnote{\cite{marquand2014bayesian} extends this approach so
    that this assumption is not strictly necessary.}.
\end{itemize}

In this work, we propose a formal definition of the problem of
decoding across subjects, and we claim that it is an instance of
transfer learning, and more precisely of \emph{transductive transfer
  learning} (TTL). Transfer learning aims at transferring knowledge
from the train set to the test set, assuming they differ in some
aspects. \emph{Transductive} transfer learning is focused on problems
where class-labels are not available for the test set, but the
unlabelled set of recordings of the test set can be used to transfer
knowledge acquired from the (class-labelled) train set. As a further
contribution, we present a simple and practical TTL solution based on
the covariate shift assumption. A final contribution is the use of the
ensemble learning principle to enhance the training process of the
classifier on the different datasets of the training subjects. We
propose to use an ensemble learning technique called \emph{stacked
  generalization} and motivate why ensemble learning is useful for our
setting. The combination of TTL and ensemble learning is the method
that we propose for decoding across subject. We show experimental
evidence of the efficacy of stacked generalization and simple
covariate shift on a face vs. scramble multisubject MEG dataset.

The article is structured as follows: in Section~\ref{sec:methods}, we
formally describe the problem of decoding across subjects and
illustrate the building blocks of the proposed approach: transfer
learning, transductive transfer learning, covariate shift and stacked
generalization. In Section~\ref{sec:experiments}, we show experimental
evidence of the efficacy of the proposed method on real MEG data and
multiple subjects. In Section~\ref{sec:discussion}, we discuss our
results and draw conclusions.


\section{Methods}
\label{sec:methods}
In this section, we formally describe decoding across subjects as a
transfer learning problem and more specifically as a transductive
transfer learning problem. Then we introduce the necessary building
blocks of the proposed solution, i.e. a simple TTL algorithm based on
the covariate shift assumption and the stacked generalization
technique of ensemble learning.

In the following, we assume that MEG data are already preprocessed.
We briefly summarise our standard pre-processing steps in
Section~\ref{sec:experiments} when discussing experiments.

\subsection{Transfer Learning}
\label{sec:transfer_learning}
Here we introduce the basic concepts of transfer learning with focus
on our application of decoding across subjects. Let $X \in
\mathcal{X}$ be the MEG recording of a single trial $(X,y)$, where $y
\in \mathcal{Y}$ represents the stimulus category presented to the
subject. In our case $\mathcal{X} = \mathbb{R}^{dC}$, where $C$ is the
number of channels of the MEG system and $d$ is the number of
timesteps recorded. Moreover, we assume a binary stimulus,
i.e. $\mathcal{Y}=\{0,1\}$. Let $P(X)$ be the marginal probability
distribution of $X$. Following~\cite{pan2010survey} we call
$\mathcal{D}=\{\mathcal{X},P(X)\}$ a \emph{domain}, which in our case
is a given subject from which we record MEG data. Given a domain, let
$f:\mathcal{X} \mapsto \mathcal{Y}$ be the predictive target function
that can be approximated from observed data. Then, for a given domain
$\mathcal{D}$, a \emph{task} is defined as a predictive function and
its output space, i.e. $\mathcal{T} = \{\mathcal{Y},f\}$. In our case,
a task could be discriminating trials when the visual stimuli are face
vs. scrambled face, or another task could be or face vs. house.

We assume to have trials recorded from a source domain $S$ and from a
target domain $T$. In our application, this means that we have trials
recorded from at least two different subjects. Moreover, in order to
provide a general definition of transfer learning, we assume that we
have a source task $\mathcal{T}_S$ and a target task
$\mathcal{T}_T$. Following the example above, these two tasks could be
face vs. scrambled face and face vs. house. Then, as stated
in~\cite{pan2010survey}, ``transfer learning aims to help improve the
learning of the target predictive function $f_T$ in $\mathcal{D}_T$
using the knowledge in $\mathcal{D}_S$ and $\mathcal{T}_S$, where, in
general, $\mathcal{D}_S \neq \mathcal{D}_T$ and $\mathcal{T}_S \neq
\mathcal{T}_T$''.

Transfer learning is a general area of research that contains
different settings, each related to a specific sub-field of machine
learning. Traditional machine learning is the case in which
$\mathcal{D}_S = \mathcal{D}_T$ and $\mathcal{T}_S =
\mathcal{T}_T$. In~\cite{pan2010survey} a taxonomy of all the settings
is illustrated and it is shown that main differences between them
depend on: how much domains and tasks differ from source to target,
and whetheror not class labels are available in the source domain or
in the target domain. The specific case of interest for this work is
defined as follows: we have MEG data from two different subjects on
the same decoding task. We use the first to train a classifier and the
second to test it, i.e. class-labels are not available on the second
subject. In other words, our problem has two different but related
domains, i.e. the two subjects, and the tasks is the same in both
cases, e.g. decoding face vs. scrambled face. Moreover class-labels
are available in the source domain but not in the target domain. This
specific setting of the transfer learning problem is called
\emph{transductive transfer learning}.

\subsection{Transductive Transfer Learning (TTL)}
\label{sec:ttl}
According to~\cite{pan2010survey}, transductive transfer learning
(TTL) aims to improve the learning of $f_T$ in $\mathcal{D}_T$ when
$\mathcal{D}_S \neq \mathcal{D}_T$ and $\mathcal{T}_S = \mathcal{T}_T$
and when unlabeled data from the target domain are available at
training time. The term \emph{transductive} stresses both the
identical task and the availability of unlabelled data.

The TTL setting can be divided into two categories:
\begin{enumerate}
\item $\mathcal{X}_S \neq \mathcal{X}_T$, i.e. source and target have
  different features spaces.
\item $P(X_s) \neq P(X_T)$, i.e. source and target share the same
  feature space, but the respective marginal probability distributions
  differ.
\end{enumerate}
Most TTL approaches available in the literature belong to the second
category. They go under the names of \emph{covariate shift} (see for
example~\cite{shimodaira2000improving}), \emph{domain adaptation} (see
for example~\cite{daume2006domain}) or \emph{sample selection
  bias}(see for example~\cite{zadrozny2004learning}). To the best of
our knowledge, transferring knowledge between different
representations, i.e. the first category, has no general solutions but
only a few domain-specific ones in domains different from
ours. See~\cite{pan2010survey} for a brief review.

In this work, we assume that $\mathcal{X}_S = \mathcal{X}_T$ and that
the difference between the data recorded from the source and target
subjects during the same task is expressed just by $P(X_s) \neq
P(X_T)$. This assumption is both meaningful and convenient. It is
meaningful because, as long as the training subject is similar to the
test subject, direct learning transfer can happen. It is also
convenient because there are some simple solutions already available
in the literature. In the following we describe the simplest one,
described by multiple authors, which is based on the idea of
importance sampling. See for
example~\cite{zadrozny2004learning,shimodaira2000improving}.

\subsubsection{Simple Covariate Shift}
\label{sec:covariate_shift}
In the empirical risk minimization framework, learning is the process
of minimizing the loss function over train data:
\begin{equation}
  \label{eq:erm}
  f^* = \argmin_{f \in \mathcal{F}} \frac{1}{n} \sum_{i=1}^n l(f, x_i,
  y_i)
\end{equation}
where $l()$ is a loss function when learning a function $f \in
\mathcal{F}$ from a dataset $\{(x_1,y_1),\ldots,(x_n,y_n)\}$. If the
train dataset is drawn from $P_S(X,Y)$ but we are interested in
predictions when the test data come from $P_T(X,Y)$, then each term
can be penalized according to how likely each trial belongs to the
target domain:
\begin{equation}
  \label{eq:erm_penalized}
  f^* = \argmin_{f \in \mathcal{F}} \frac{1}{n} \sum_{i=1}^n
  \frac{P_T(x_i,y_i)}{P_S(x_i,y_i)} l(f, x_i, y_i).
\end{equation}
The covariate shift assumption is that $P_S(Y|X) = P_T(Y|X)$, then
$\frac{P_T(x_i,y_i)}{P_S(x_i,y_i)} = \frac{P_T(x_i)}{P_S(x_i)}$. Then,
the risk minimization problem becomes:
\begin{equation}
  \label{eq:erm_penalized}
  f^* = \argmin_{f \in \mathcal{F}} \frac{1}{n} \sum_{i=1}^n
  \frac{P_T(x_i)}{P_S(x_i)} l(f, x_i, y_i).
\end{equation}
Among the various ways to estimate $\frac{P_T(x_i)}{P_S(x_i)}$, a
simple one is to set up a new classification problem to discriminate
trials belonging to the source domain from those of the target domain,
which requires just unlabeled data.

\subsection{Ensemble Learning: Stacked Generalization}
\label{sec:stacking}
It is common experience that, when single-subject decoding is
accurate, decoding across subject, by just pooling all trials of train
subjects together, is not accurate on trials from test subjects. As
previously discussed, the cause of this issue is the variability
across subjects. This situation shares analogy with \emph{ensemble
  learning}, where multiple classifiers are trained from portions of
the train set and then combined to create a single prediction on test
data. The underlying ideas of ensemble learning are mainly two: first,
each classifier in the ensemble tries to capture specific aspects of
the train set. Second, the additional step of combining the
classifiers of the ensemble tries to reduce the instability of the
prediction of each of them. In short, diversity of classifiers and
stability of the combined predictions are the two key elements of
ensemble learning. In our problem of decoding across subjects, the set
of trials of the train subjects is naturally partitioned in a way that
training one classifier for each subject is expected to well represent
the diversity within the train data. For this reason, we claim that
combining these classifiers can be more effective than ignoring the
differences and pooling all subjects.

In the vast literature of ensemble learning, multiple approaches have
been proposed, such as bagging, boosting, Bayesian model averaging and
stacked generalization. A reference book for this topic
is~\cite{kuncheva2004combining}. Among the many techniques available
in this field of research, here we propose to use \emph{stacked
  generalization}, also known as \emph{stacking} classifiers, which
aims at learning the combination of
classifiers. See~\cite{wolpert1992stacked,ting1999issues} for a
detailed introduction to stacked generalization. Here, we briefly
introduce the basic procedure in the context of the problem of
decoding across subjects.

The procedure of stacked generalization is divided in the following
steps:
\begin{enumerate}
\item Train a set of classifiers on (portions of) the train
  data. These classifiers are called \emph{first-level} classifiers.
\item Collect the output of each classifier on each trial of the train
  and of the test data. These outputs are called first level
  predictions.
\item Create a new \emph{second-level} dataset with the vector of
  first-level predictions for each trial. Care has to be taken so that
  the predicted value of a given trial comes from classifiers which
  were not trained on that trial, e.g. through cross-validation.
\item The class-labels of the second-level dataset are the same as the
  initial dataset.
\item A second-level classifier is trained on the portion of the
  second-level dataset related to the train subjects in order to learn
  how to combine the first-level predictions.
\item The second level classifier is used to predict the class-labels
  of the test data as represented in the second-level dataset.
\end{enumerate}
In this work, we claim that creating each first-level classifier on
the data of one subject only, is an effective way to ensure diversity
which is a key element for the success of ensemble learning. Moreover,
this approach implicitly assumes that the test set can be represented
as a combination of the different patterns observed in the data of the
training subjects, which is a desirable property for our specific
problem.

As a final step, we propose to combine covariate shift and stacked
generalization in the following way: once the second level classifier
is created, each instance related to the training subjects is weighted
according to the simple re-weighting procedure described
in~\ref{sec:covariate_shift}.




\section{Experiments}
\label{sec:experiments}
We tested the proposed method on an MEG dataset where subjects were
presented visual stimuli about famous faces, unfamiliar faces and
scrambled faces. The dataset is from a multi-modal study described
in~\cite{henson2011parametric} and consists of $16$ subjects. We refer
the reader to the original study for a comprehensive description of
the dataset. We created a balanced face vs. scramble dataset by
drawing at random from the trials of famous and unfamiliar faces in
equal number to that scrambled faces. The dataset of each subject
consisted of $\approx 580$ trials, so the entire dataset consisted of
9414 trials. We high-pass filtered the raw data at 1Hz, downsampled to
200Hz, and epoched each trial to the first 500ms after stimulus
onset. Then each trial consisted of 100 timepoints for each of the 306
MEG channels, which we concatenated into a vector of $100 \times 306 =
30600$ elements that was the input of the classification algorithms.

First, we attempted single-subject decoding, using logistic regression
with $\ell_1$ penalization. The column ``single'' in
Table~\ref{tab:results} shows the 6-fold cross-validated (CV)
accuracies. The results show that high accuracy could be reached in
all cases, with an average accuracy of $0.82$ and a range of
$0.70-0.90$.

We conducted a leave-one-subject out cross-validation with the three
different methods described in Section~\ref{sec:methods}. The results
are reported in Table~\ref{tab:results} and the corresponding columns
are indicated in the following. The first method (column ``pool'')
consisted in just pooling all trials of all subjects and using
logistic regression with $\ell_1$ penalization. We observe that the
average accuracy drops to $0.62$, which is a $0.2$ reduction with
respect to the single-subject decoding accuracy.

The second method (column ``SG'' in Table~\ref{tab:results}) is based
on stacked generalization where both the first-level and second-level
classifiers were again logistic regression. The average accuracy is
$0.65$ and consistently greater that pooling over almost all
subjects. The third method (column ``SG+CS'' in
Table~\ref{tab:results}) is the same as the second method but
additionally each trial of the second-level dataset was weighted with
a coefficient computed according to the simple covariate-shift
technique described in Section~\ref{sec:covariate_shift}. The
classifier used to compute the weight of each trial was again logistic
regression. The average accuracy is $0.67$, and is consistently
greater or equal to that of plain stacked generalization over almost
all subjects.

\begin{table}
  \centering
  \begin{tabular}{c | c || c | c | c |}
    Subj. & single & pool & SG & SG+CS \\
    \hline
    1   & 0.82 & 0.62 & 0.67 & \textbf{0.71} \\
    2   & 0.75 & 0.64 & 0.63 & \textbf{0.65} \\
    3   & 0.82 & 0.60 & 0.59 & \textbf{0.61} \\
    4   & 0.86 & 0.70 & \textbf{0.75} & 0.72 \\
    5   & 0.80 & 0.58 & 0.63 & \textbf{0.69} \\
    6   & 0.81 & \textbf{0.65} & 0.60 & 0.60 \\
    7   & 0.82 & 0.61 & 0.68 & \textbf{0.72} \\
    8   & 0.85 & 0.64 & 0.66 & \textbf{0.71} \\
    9   & 0.87 & 0.67 & 0.71 & \textbf{0.73} \\
    10  & 0.81 & 0.57 & 0.66 & \textbf{0.70} \\
    11  & 0.70 & 0.60 & 0.65 & \textbf{0.67} \\
    12  & 0.78 & 0.65 & \textbf{0.67} & 0.66 \\
    13  & 0.88 & 0.60 & 0.64 & \textbf{0.66} \\
    14  & 0.86 & 0.69 & \textbf{0.68} & \textbf{0.68} \\
    15  & 0.90 & 0.67 & 0.67 & \textbf{0.70} \\
    16  & 0.87 & 0.52 & \textbf{0.57} & 0.56 \\
    \hline
    \hline
    mean & 0.82 & 0.62 & 0.65 & \textbf{0.67}
  \end{tabular}
  \caption{Classification accuracies across 16 subjects in a face
    vs. scrambled visual task. The column ``single'' reports the
    single-subject decoding accuracies (6 fold CV). The last three
    columns show leave-one-subject-out accuracies for pooling (pool),
    stacked generalization (SG) and stacked generalization with
    covariate shift (SG+CS). The last row shows the mean accuracies
    across subjects.}
  \label{tab:results}
\end{table}


\section{Discussion and Conclusion}
\label{sec:discussion}
In this paper, we formally presented the problem of MEG decoding
across subjects for inferential purpose. We positioned the problem
with respect to the current literature and in particular within the
framework of transductive transfer learning (TTL). We presented a
basic approach to TTL, i.e. simple covariate shift, based on trial
weighting in order to penalize the training trials far from the test
trials.

We draw an analogy between the problem of decoding across subjects and
ensemble learning. We introduce stacked generalization as an
ensembling method to deal the variability across the triaining
subjects. The final method we proposed combines stacked generalization
(SG) and covariate shift (CS).

The experiments presented in Section~\ref{sec:experiments} compare the
proposed approach to the baseline approach of simply pooling all the
trials of all the training subjects without attempting to model their
differences. The results in Table~\ref{tab:results} show that both SG
and CS are able to extract information from the similarities and
differences between subjects. The baseline approach has average
accuracy $0.62$ while SG alone reaches $0.65$ and SG together with CS
reach $0.67$.

The results show a large decrease in accuracy between decoding across
subjects and decoding single subjects, where accuracy reaches $0.82$
on average. This gap motivates the need for further research.


\bibliographystyle{plain}

\bibliography{emanueleolivetti-prni2014_across}

\end{document}